\documentclass[conference]{IEEEtran}

\usepackage{cite}
\usepackage{amsmath,amssymb,amsfonts}
\usepackage{algorithmic}
\usepackage{graphicx}
\usepackage{textcomp}
\usepackage{xcolor}
\usepackage{url}
\usepackage{booktabs}
\usepackage{multirow} 
\usepackage{cleveref}
\usepackage{tipa}

\def\BibTeX{{\rm B\kern-.05em{\sc i\kern-.025em b}\kern-.08em
    T\kern-.1667em\lower.7ex\hbox{E}\kern-.125emX}}
\begin{document}

\title{Quantifying Consonant Contributions to \\ Word Intelligibility via Acoustic Masking}


\author{\IEEEauthorblockN{Eunjung Yeo\IEEEauthorrefmark{2}, Kwanghee Choi\IEEEauthorrefmark{2}, 
Krupaben Kothadia\IEEEauthorrefmark{4}, \\
Visar Berisha\IEEEauthorrefmark{4},
Julie M. Liss\IEEEauthorrefmark{4}, 
David R. Mortensen\IEEEauthorrefmark{5}, 
David Harwath\IEEEauthorrefmark{2}}
\IEEEauthorblockA{\IEEEauthorrefmark{2}University of Texas at Austin, USA, \IEEEauthorrefmark{4}Arizona State University, USA, \IEEEauthorrefmark{5}Carnegie Mellon University, USA} 
\{eunjung.yeo,harwath\}@utexas.edu}

\maketitle

\begin{abstract} 
Consonants contribute unequally to whether a word is understood. Given the limited time available for therapy, ranking consonants by contribution to intelligibility helps prioritize intervention targets in motor speech disorders. However, measuring this contribution relies on perceptual studies that are difficult to scale. This paper presents a scalable method that measures consonant contribution using acoustic masking. We silence one consonant at a time in an isolated word and test whether an automatic speech recognition (ASR) model still recognizes the word. We define a consonant's contribution score as the proportion of its masked instances for which the word becomes misrecognized, which we refer to as the mask-induced misrecognition rate (MMR). We relate MMR to two linguistic factors previously reported to correlate with consonant contribution, namely phoneme frequency and functional load. We apply this analysis across four languages, English, Spanish, German, and Czech, using three ASR architectures, MMS (encoder-only), Whisper (encoder-decoder), and Qwen3-ASR (LLM-based). Using partial Spearman correlations, we find that phoneme frequency correlates negatively with MMR while functional load correlates positively. In other words, more frequent consonants are less disruptive when masked, whereas consonants carrying more lexical contrast are more disruptive. Further cross-language analysis shows that consonant rankings agree only partially across languages, indicating that consonant contribution is language-dependent.
\end{abstract}



\section{Introduction}\label{sec:introduction}
Speech intelligibility depends unevenly on individual phonemes \cite{millernicely1955, kent1989phonetic}. Certain phonemes, when mispronounced, cause a listener to misidentify a word, while others have limited effect on intelligibility \cite{kim2017cross, yeo2023speech}. Knowing how much each phoneme contributes to intelligibility has practical consequences wherever pronunciation accuracy is a concern, from second-language learners \cite{munro1995foreign} to individuals with motor speech disorders \cite{hustad2006estimating, lousada2014intelligibility}, where limited intervention time makes prioritization especially valuable.


In motor speech disorders (MSD), imprecise articulation is among the strongest contributors to intelligibility loss \cite{de2002intelligibility}. Phoneme-level articulation therapy is therefore a common intervention, with programs such as boost articulation therapy (BArT) drilling the phonemes a speaker misarticulates \cite{mendoza2021effect}. Which phonemes to prioritize, though, is left to the clinical judgment of Speech-Language Pathologists (SLPs) \cite{asha_dysarthria_adults}, informed by several criteria \cite{deveney2020target}. These include stimulability, to target sounds the speaker can approximate with guidance, ease of articulation, to favor simpler motor demands, and phoneme frequency, to prioritize sounds that recur in everyday speech.

Alongside these criteria, a phoneme's contribution to intelligibility offers another consideration for prioritization, that is, how much its correct production determines whether a word is understood. A phoneme can be stimulable, easy to articulate, or frequent, yet still contribute little to whether a word is recognized. Adding contribution as a further dimension could help direct limited therapy time toward the sounds that contribute most to intelligibility among the mispronounced phonemes. The functional importance to intelligibility (FITI) framework \cite{gurevich2023, gurevichkim2024, gurevich2024frequency} was recently proposed to capture this dimension, attributing a phoneme's contribution to linguistic factors such as positional prominence, frequency, and functional load. While the framework is intended to apply across languages, it has been perceptually validated only in English, and only for a small set of consonants \cite{gurevich2024frequency}.

\begin{figure*}[t!]
    \centering
    \includegraphics[width=0.95\textwidth]{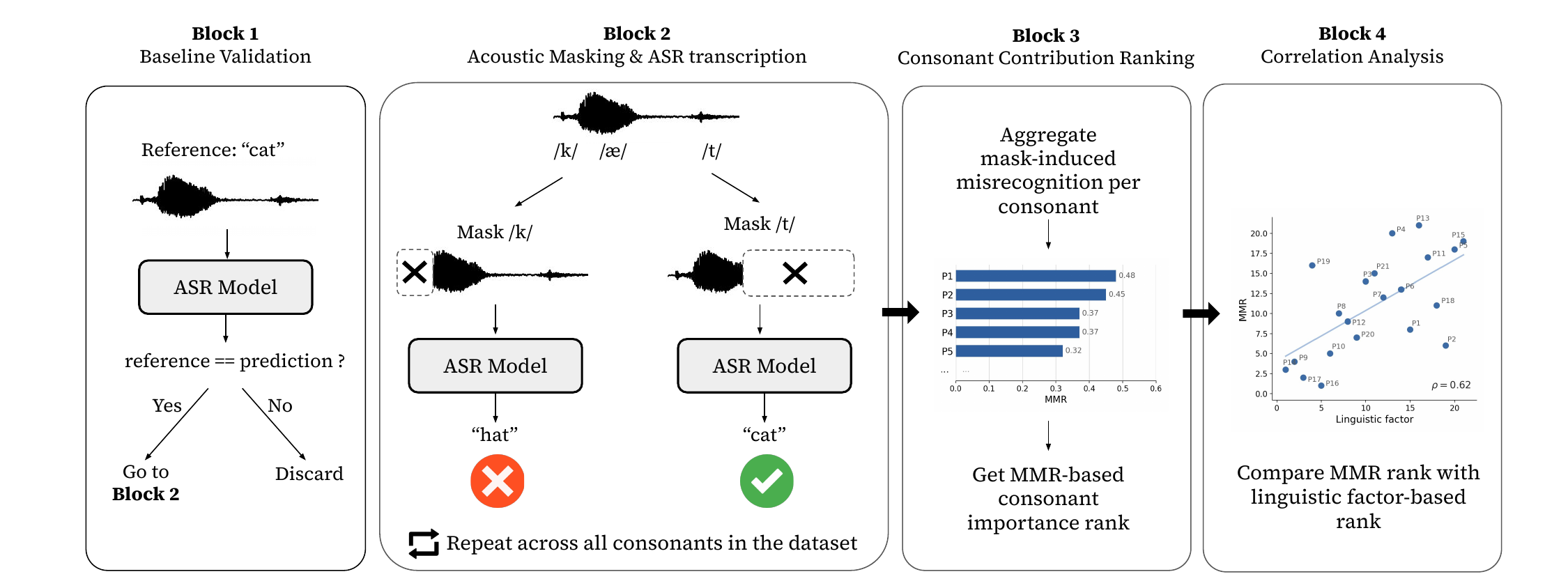}
    \caption{Overview of the experimental pipeline. Block 1 (Baseline Validation): words are transcribed by the ASR model. Only correctly recognized words are retained for masking. Block 2 (Acoustic Masking \& ASR transcription): each target consonant is silenced and the masked word is re-transcribed. A mismatch between reference and prediction is recorded as a misrecognition. Block 3 (Consonant Contribution Ranking): for each consonant, the fraction of masking trials that cause misrecognition gives its mask-induced misrecognition rate (\texttt{MMR}). Block 4 (Correlation Analysis): Rank of \texttt{MMR}s and two linguistic factors are compared.}\label{fig:overview}
\end{figure*}

Measuring phoneme contribution directly is difficult. Contribution to intelligibility is ultimately defined by human perception, so measuring it requires listener judgments for every phoneme in every position. This difficulty compounds across languages. Because contribution is shaped by language-specific phonological structure \cite{kim2017cross, yeo2025crosslang}, each language would require its own perceptual study with native listeners and language-specific stimuli. Conducting such studies for every language of interest is prohibitively costly.

Automatic speech recognition (ASR) offers a scalable alternative to perceptual evaluation. ASR-derived word error rate (WER) has become a standard proxy for human intelligibility judgments of synthesized speech from TTS and voice conversion \cite{taylor2021, alharthi2024}. WER has also been reported to correlate reliably with clinician-rated intelligibility for dysarthric speakers \cite{karbasi2022asr, choi2026automatic}. Because ASR runs cheaply on large corpora, it can measure phonemes across the natural distribution of contexts and words in a language, coverage that perceptual studies cannot afford. Moreover, recent multilingual ASR models let the WER-based measurement extend to many languages.

This paper presents a scalable method to quantify how much individual consonants contribute to word intelligibility across languages. We focus on consonants because they constrain lexical identification more tightly than vowels \cite{cutler2000constraints}. For each consonant, we mask its acoustic interval in an isolated word and test whether an ASR model still recognizes the word. We define each consonant's contribution as the mask-induced misrecognition rate (MMR), the proportion of masked instances for which the word is misrecognized. We calculate MMR on typical speech to obtain a language-level prior, independent of any individual speaker's impairments. We evaluate on English, Spanish, German, and Czech, where the latter three are
the focused-track languages of the \textit{OneVoice-MSD 2026} ~\cite{hernandez2026adapting}.

In summary, our contributions are as follows:
\begin{itemize}
    \item We introduce the \textbf{mask-induced misrecognition rate (MMR)}, a language-agnostic method that quantifies consonant contribution to word intelligibility through ASR-based acoustic masking, applicable to any language with a forced aligner and a multilingual ASR model.
    \item We relate MMR to linguistic factors for consonant contribution: phoneme frequency and functional load.
    \item Applying MMR to English, Spanish, German, and Czech, we find that consonant rankings agree only partially across languages, indicating that consonant contribution is language-dependent.
\end{itemize}

\section{Related work}
\textbf{Phoneme contribution to intelligibility.}
Phonemes differ in how much they contribute to intelligibility \cite{kent1989phonetic, anselkent1992, owren2006relative}. For example, consonants carry more lexical information than vowels in isolated words \cite{cutler2000constraints}, and pronunciation accuracy on certain phonemes correlates with intelligibility more than on others \cite{yeo2023speech}. \textit{Ranking} individual consonants by their contribution to intelligibility, however, has received less attention. The FITI framework \cite{gurevich2023, gurevichkim2024, gurevich2024frequency} offers a conceptual hierarchy, deriving consonant importance jointly from linguistic factors such as frequency, functional load, and positional prominence. Its validation so far covers two English consonants, \textipa{/\textturnr/} and \textipa{/\texttheta/} \cite{gurevich2024frequency}. Building on this foundation, we use ASR to provide empirical measures of consonant contribution across the full consonant inventory and four languages.

\textbf{Masking as a probe.}
Measuring how recognition degrades when part of the signal is removed or perturbed is an established probe of which segments carry intelligibility \cite{millernicely1955, fogerty2012relative}. If the utterance is correctly recognized after masking, the removed segment was recoverable from the rest of the input. If not, the segment carried information the rest of the input did not supply. Recent work automates this with ASR, adding noise to phonemes in full sentences and reading intelligibility from the recognizer output \cite{drelingyte2025}. The authors note that word-level degradation partly reflects the recognizer's use of sentence context rather than the masked segment alone. We therefore decode each word in isolation, so the probe isolates recoverability from within the word rather than from surrounding context. This parallels established single-word approaches to intelligibility testing in dysarthria \cite{yorkston1981aids, kent1989phonetic}.

\section{Method}\label{sec:method}
\Cref{fig:overview} summarizes our experimental pipeline. We replace a consonant's interval with silence, leaving the remainder of the signal intact. As silence eliminates all acoustic information within the interval, whereas real mispronunciations often retain partial cues such as manner of articulation or spectral shape, the resulting measure constitutes an upper bound on the consonant's contribution.

\textbf{1. Forced alignment.}
    Each utterance is force-aligned to its transcription at the word and phoneme levels, identifying the interval $[p_\text{start}, p_\text{end}]$ of every phoneme together with the boundaries of its enclosing word $[w_\text{start}, w_\text{end}]$. 

\textbf{2. Baseline transcription.} 
    Each word is cropped to $[w_\text{start}, w_\text{end}]$ and transcribed in isolation by the ASR model (\Cref{ssec:asr-model}). Only words correctly transcribed are retained, ensuring that any subsequent misrecognition is attributable to masking rather than a pre-existing error. Homophones are treated as correct transcriptions (\textit{e.g.}, \textit{write} and \textit{right}).

    \textbf{3. Silence masking and re-transcription.} For each word, each consonant interval is masked with silence, one at a time, with a margin to suppress coarticulatory leakage from neighboring segments (\Cref{ssec:silence-masking}). Each masked word is then re-transcribed independently by the ASR model.

    \textbf{4. Scoring and aggregation.}
    Re-transcription is scored as a mask-induced misrecognition if it no longer matches the reference word, and correct otherwise, again treating homophones as correct. For each consonant, we average over all instances to obtain its contribution score to intelligibility, \textit{i.e.}, \textbf{mask-induced misrecognition rate (\texttt{MMR})}. The scores are computed separately per language and ASR model. 


\section{Experimental Setup}\label{sec:setup}
\subsection{Datasets}\label{sec:dataset} 
We use five corpora spanning four languages: English, Spanish, German, and Czech, where the latter three are the focused-track languages of the \textit{OneVoice-MSD 2026} \cite{hernandez2026adapting}.

\textbf{Common Voice (CV)} \cite{ardila2020common} is a large-scale crowdsourced read-speech corpus with sentence-level transcripts spanning multiple languages, providing broad phonemic coverage over commonly used vocabulary. We use version 22.0, with the test set for acoustic masking (\Cref{sec:alignment-masking}) and the train and development sets for estimating the linguistic factors (\Cref{{ssec:predictor-calc}}).

\textbf{TIMIT} \cite{garofolo1993timit} is an acoustic-phonetic corpus of American English with hand-verified, time-aligned phone boundaries. We use TIMIT as the second English dataset, masking its test set and comparing the results against linguistic factors estimated from Common Voice English. This lets us test whether the link between \texttt{MMR} and the linguistic factors holds across independent corpora of the same language.

\subsection{Acoustic masking} \label{sec:alignment-masking}
\subsubsection{Forced alignment} 
For TIMIT, we use the dataset's manual phone segmentation: ARPABET labels are lowercased and de-stressed, stop closures are merged with their following burst (\textit{e.g.}, [pcl][p]$\rightarrow$[p]), glottal stops and silences are discarded. For Common Voice, we automatically align using the Montreal Forced Aligner (MFA) \cite{mcauliffe2017montreal}. Out-of-vocabulary words are transcribed using Epitran \cite{mortensen2018epitran} Grapheme-to-Phoneme (G2P) model\footnote{Language codes for Epitran: English (\texttt{eng-Latn}), Spanish (\texttt{spa-Latn-eu}), German (\texttt{deu-Latn-np}), and Czech (\texttt{ces-Latn}).}, and added to the pronunciation dictionary prior to forced alignment. All audio is processed at 16kHz. 

\subsubsection{Silence masking} \label{ssec:silence-masking}
For each isolated word, each consonant is masked by replacing its forced-aligned interval with silence. To suppress acoustic leakage from neighboring phonemes due to coarticulation, the silence window is extended by 25\% of the duration of each adjacent phoneme. For the word-initial and word-final phonemes, the window is extended only on one side, clipped to the word boundary. 

\subsubsection{ASR transcription}\label{ssec:asr-model}
We use three multilingual ASR models spanning the main architectural families in current speech recognition: encoder-only (MMS), encoder-decoder (Whisper), and LLM-based (Qwen3-ASR), all supporting the four languages under analysis. All experiments were run on a single NVIDIA Quadro RTX 8000 with a batch size of 32.

\begin{figure*}[t!]
    \centering
    \includegraphics[width=\textwidth]{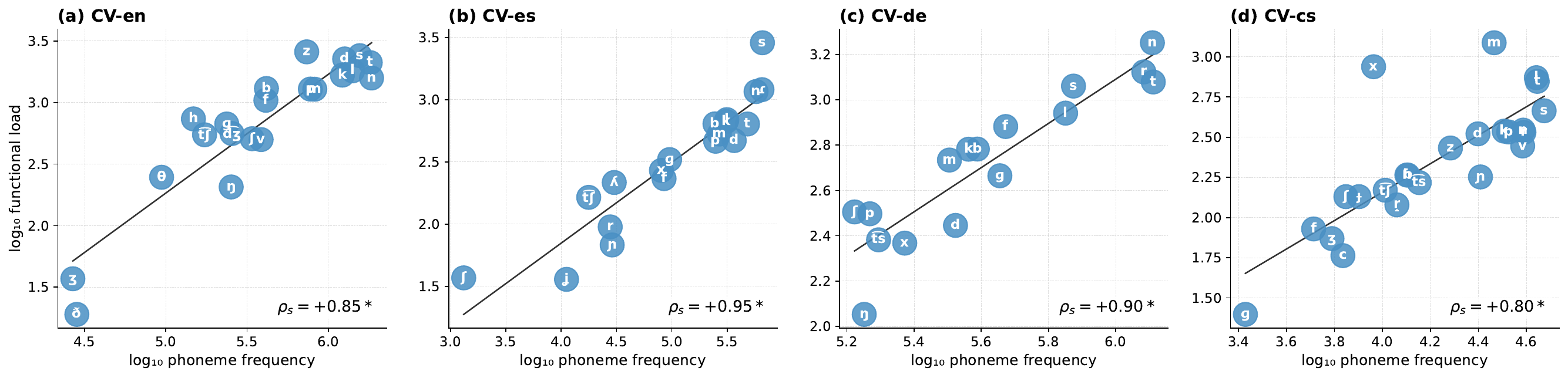}
\caption{Phoneme frequency (\texttt{freq}) and functional load (\texttt{FL}) are positively correlated across all four languages (log–log axes).}
\label{fig:collinearity}
\end{figure*} 

\textbf{MMS}~\cite{pratap2024scaling} is an encoder-only model self-supervised on $\sim$500K hours of speech across 1{,}400+ languages, then fine-tuned for ASR with CTC loss \cite{graves2006connectionist} on 44.7K hours spanning 1{,}107 languages. We use \texttt{facebook/mms-1b-all} with its per-language CTC head and language adapter.

\textbf{Whisper}~\cite{radford2023robust} is an encoder-decoder model trained on 680K hours of weakly supervised, web-crawled audio across 99 languages. We use \texttt{openai/whisper-large-v3-turbo}, supplying the target language token to condition the decoder.

\textbf{Qwen3-ASR}~\cite{Qwen3-ASR} is an LLM-based ASR model that couples the audio encoder with the Qwen3-Omni LLM~\cite{xu2025qwen3}, acquiring ASR through post-training. Its audio encoder is pretrained on $\sim$40M hours of pseudo-labeled speech, and the released model spans 52 languages. We use \texttt{Qwen/Qwen3-ASR-1.7B}, with the target language passed at inference time.

\textbf{Trial counts.} 
\Cref{tab:dataset} reports the number of word tokens retained for the acoustic masking experiment. The number of retained tokens follows $\text{MMS} < \text{Whisper} < \text{Qwen}$ in all corpora except Czech. Qwen drops below Whisper for Czech, consistent with the Qwen3-ASR report's acknowledgement of weaker performance on lower-resource languages compared to Whisper \cite{Qwen3-ASR}.

\begin{table}[t]
\centering
\small
\setlength{\tabcolsep}{4pt}
\caption{Word token counts used in the study. Unique word counts in parentheses. Train+dev are used for linguistic factor estimation while baseline-correct tokens are used as trials for the acoustic masking experiment.}\label{tab:dataset}
\resizebox{\columnwidth}{!}{%
\begin{tabular}{l l r r r r}
\toprule
 & & & \multicolumn{3}{c}{Baseline-correct} \\
\cmidrule(lr){4-6}
Dataset & Language & Train+dev & MMS & Whisper & Qwen \\
\midrule
\multirow{4}{*}{CV \cite{ardila2020common}}
 & English (en) & 4{,}385k (12.3k) & 13.3k & 29.6k & 34.5k \\
 & Spanish (es) & 1{,}183k (11.7k) & 23.6k & 32.1k & 34.6k \\
 & German  (de) & 2{,}059k (14.1k) & 18.0k & 29.3k & 33.0k \\
 & Czech   (cs) &    133k (18.9k)  & 11.1k & 11.6k & 10.2k \\
\midrule
TIMIT \cite{garofolo1993timit}
 & English (en) & ---              &  1.8k &  3.5k &  3.8k \\
\bottomrule
\end{tabular}
}
\end{table}

\subsection{Linguistic factors for phoneme contributions}\label{ssec:predictor-calc}
We correlate \texttt{MMR} with two linguistic factors, phoneme frequency and functional load, to test whether it aligns with established measures of consonant contribution. Both factors can be computed for any phoneme, but we restrict our analysis to consonants, consistent with the scope of this paper.

\subsubsection{Lexicon preparation} 
Both linguistic factors are computed from the reference transcriptions of Common Voice train+dev set. We first tokenize each transcript by lowercasing and splitting on whitespace and punctuation. We then apply Epitran \cite{mortensen2018epitran} for G2P conversion, whose output is segmented into phonemes with PanPhon \cite{mortensen2016panphon}. We restrict the lexicon to words occurring at $\geq$\,10 per million tokens to filter out uncommon words in the language.

\subsubsection{Phoneme Frequency (\texttt{freq})} 
\texttt{freq} reflects how often a phoneme occurs in the language. More frequent phonemes are more predictable across the word
contexts in which they occur, and may therefore remain
recoverable even when acoustically degraded. For each phoneme $p$, we sum its token-weighted occurrences across the lexicon,
\begin{equation}
    c(p) = \sum_{w} n_w \, \#_p(w),
\end{equation}
where $n_w$ is the token count of word $w$ and $\#_p(w)$ is the number of times $p$ appears in its phoneme sequence. We use $\log_{10} c(p)$ as the frequency measure.

\subsubsection{Functional Load (\texttt{FL})}
\texttt{FL} quantifies the communicative work a phoneme performs in maintaining lexical distinctions. Phonemes with high functional load distinguish many word pairs, so their loss would produce more lexical ambiguity. Functional load has been defined in different ways, including minimal pair counts \cite{martinet1952function, hockett1967quantification}, frequency-weighted minimal pair counts \cite{wedel2013high}, and entropy-based measures \cite{surendran2003quantitative}. Following FITI \cite{gurevich2023}, we adopt the unweighted minimal pair count. For phoneme $p$, we count the minimal pairs in which $p$ is the differing sound. These pairs differ by either a substitution (\textit{e.g.}, \textit{cat} vs. \textit{bat}) or a single insertion or deletion (\textit{e.g.}, \textit{block} vs. \textit{lock}). We use $\log_{10}$ values as the functional load measure.


\subsection{Word filtering}\label{ssec:word-filtering}
Both acoustic masking (\Cref{sec:alignment-masking}) and linguistic factor calculations (\Cref{ssec:predictor-calc}) operate on the content words of each language. Following the FITI framework \cite{gurevich2023, gurevichkim2024}, function words are excluded because they are largely recoverable from grammatical context, so their recognition depends less on the acoustic realization of their segments. We approximate the content word set by removing words listed in language-specific stopword lists \cite{stopwordsiso}.

\subsection{Evaluation}\label{ssec:evaluation}
\textbf{Phone-to-Phoneme Mapping.}
Before aggregating mask-induced misrecognition per phoneme (\Cref{fig:overview}, Block 3), we normalize MFA's surface phone labels to the canonical phoneme inventory produced by Epitran, which is the inventory used for \texttt{freq} and \texttt{FL} estimation. Surface variants such as aspirated stops, place-assimilated nasals, and final-devoiced obstruents are collapsed to their canonical underlying forms. This ensures that each consonant's \texttt{MMR} and its \texttt{freq} and \texttt{FL} are computed over the same inventory. Unmappable phones are excluded, which affects less than 10\% of tokens per dataset.

\textbf{Rank correlation.}
Spearman rank correlations $\rho$ are computed between \texttt{MMR} and each linguistic factor. As \texttt{freq} and \texttt{FL} are strongly correlated with each other, we additionally compute partial Spearman correlations to isolate the unique association of each factor with contribution while holding the other constant. 
\begin{figure*}[t!]
    \centering
    \includegraphics[width=\textwidth]{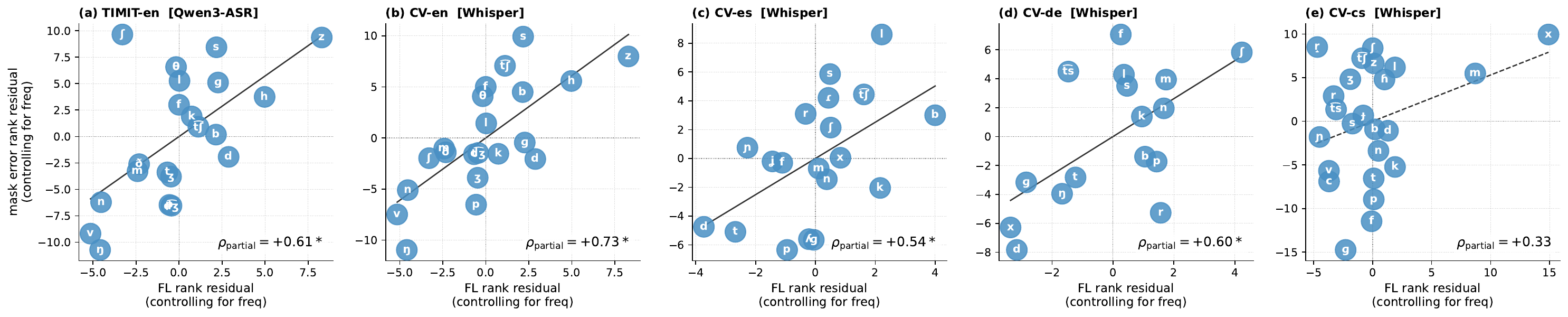}
\caption{
Phoneme-level partial correlation (\cref{eq:partial_spearman}) between mask-induced misrecognition rate (\texttt{MMR}) and functional load (\texttt{FL}), controlling for phoneme frequency (\texttt{freq}). Qwen3-ASR for TIMIT-en, Whisper for all others.
The regression line is solid when it is significant ($p < .05$) and dashed otherwise.}
\label{fig:partial-spearman}
\end{figure*} 
Partial Spearman correlation extends Spearman's $\rho$ to control for a confounding variable $Z$. Given three variables $X$, $Y$, and $Z$ (all converted to ranks), the partial correlation between $X$ and $Y$ controlling for $Z$ is: 
\begin{equation} 
\rho(X, Y | Z) = \frac{\rho(X, Y) - \rho(X, Z)  \rho(Y, Z)} {\sqrt{(1 - \rho(X, Z)^2)(1 - \rho(Y, Z)^2)}} ,\label{eq:partial_spearman} 
\end{equation} 
where $\rho(X, Y)$, $\rho(X, Z)$, and $\rho(Y, Z)$ are the pairwise Spearman correlations. Intuitively, $\rho(X, Y | Z)$ captures the monotonic association between $X$ and $Y$ that cannot be explained by their shared relationship with $Z$.

\begin{table}[t]
\centering
\caption{Spearman correlation between Functional Load (\texttt{FL}) or Phoneme Frequency (\texttt{freq}) and Mask-induced Misrecognition Rate (\texttt{MMR}). $^{*}p<.05$, $^{\dagger}p<.1$}
\resizebox{\columnwidth}{!}{%
\label{tab:fl_corr}
\begin{tabular}{l lllll}
  \toprule
  Model & TIMIT-en & CV-en & CV-es & CV-de & CV-cs \\
  \midrule
  \multicolumn{6}{l}{\emph{A. Correlation}\ $\rho(\texttt{MMR}, \texttt{freq})$} \\
    \quad MMS      & $-0.11$ & $-0.56^{*}$ & $-0.44^{\dagger}$ & $-0.27$ & $-0.20$ \\
    \quad Whisper  & $-0.03$ & $-0.49^{*}$ & $-0.65^{*}$ & $-0.22$ & $-0.28$ \\
    \quad Qwen3-ASR     & $-0.18$ & $-0.43^{\dagger}$ & $-0.58^{*}$ & $-0.14$ & $-0.40^{\dagger}$ \\
    \quad \textit{Average}     & $-0.10$ & $-0.42^{\dagger}$ & $-0.52^{*}$ & $-0.20$ & $-0.33$ \\
  \midrule
  \multicolumn{6}{l}{\emph{B. Correlation}\ $\rho(\texttt{MMR}, \texttt{FL})$} \\
    \quad MMS      & $+0.19$ & $-0.22$ & $-0.28$ & $-0.03$ & $+0.02$ \\
    \quad Whisper  & $+0.26$ & $-0.09$ & $-0.48^{*}$ & $+0.06$ & $-0.04$ \\
    \quad Qwen3-ASR     & $+0.16$ & $-0.05$ & $-0.43^{\dagger}$ & $+0.10$ & $-0.17$ \\
    \quad \textit{Average}     & $+0.22$ & $-0.02$ & $-0.35$ & $+0.05$ & $-0.10$ \\
  \midrule
  \multicolumn{6}{l}{\emph{C. Partial Correlation}\ $\rho(\texttt{MMR}, \texttt{freq}\mid\texttt{FL})$} \\
    \quad MMS      & $-0.54^{*}$ & $-0.74^{*}$ & $-0.54^{*}$ & $-0.55^{*}$ & $-0.37^{\dagger}$ \\
    \quad Whisper  & $-0.49^{*}$ & $-0.80^{*}$ & $-0.68^{*}$ & $-0.62^{*}$ & $-0.42^{*}$ \\
    \quad Qwen3-ASR     & $-0.61^{*}$ & $-0.74^{*}$ & $-0.59^{*}$ & $-0.51^{\dagger}$ & $-0.45^{*}$ \\
    \quad \textit{Average}     & $-0.56^{*}$ & $-0.78^{*}$ & $-0.62^{*}$ & $-0.55^{*}$ & $-0.42^{*}$ \\
  \midrule
  \multicolumn{6}{l}{\emph{D. Partial Correlation}\ $\rho(\texttt{MMR}, \texttt{FL}\mid\texttt{freq})$} \\
    \quad MMS      & $+0.55^{*}$ & $+0.60^{*}$ & $+0.45^{\dagger}$ & $+0.50^{\dagger}$ & $+0.31$ \\
    \quad Whisper  & $+0.54^{*}$ & $+0.73^{*}$ & $+0.54^{*}$ & $+0.60^{*}$ & $+0.33$ \\
    \quad Qwen3-ASR     & $+0.61^{*}$ & $+0.67^{*}$ & $+0.45^{\dagger}$ & $+0.51^{\dagger}$ & $+0.27$ \\
    \quad \textit{Average}     & $+0.58^{*}$ & $+0.72^{*}$ & $+0.50^{*}$ & $+0.53^{*}$ & $+0.29$ \\
  \bottomrule
\end{tabular}
}
\end{table}

\section{Results}\label{sec:results}
We examine how \texttt{freq} and \texttt{FL} rank consonants by mask-induced misrecognition rate (\texttt{MMR}).
All correlations here are computed over orderings, not values.
We report both types of Spearman correlations in \Cref{tab:fl_corr}. 

As for marginal Spearman correlation, \texttt{MMR} and \texttt{freq} are negatively correlated (Block A in \Cref{tab:fl_corr}), and \texttt{MMR} and \texttt{FL} show no consistent pattern (Block B in \Cref{tab:fl_corr}).
Yet, \texttt{FL} and \texttt{freq} are strongly correlated ($\rho = {+}0.80$ to ${+}0.95$, \Cref{fig:collinearity}), indicating that these correlations can be misleading. 

We therefore compute partial Spearman correlations to separate the two factors. Controlling for \texttt{FL}, the partial correlation between \texttt{MMR} and \texttt{freq} becomes more strongly negative (Block C). Controlling for \texttt{freq}, the partial correlation between \texttt{MMR} and \texttt{FL} becomes positive in every condition and significant in most cases (Block D). These signs indicate that more frequent consonants are less disruptive when masked, whereas consonants carrying more lexical contrast are more disruptive.
One exception is the partial correlation between \texttt{MMR} and \texttt{FL} for Czech, which stays positive while not reaching significance.
We attribute this to noisier estimates on both sides: a smaller Czech train+dev set yields noisier \texttt{freq} and \texttt{FL}, and the lower ASR performance on Czech (\Cref{ssec:asr-model}) yields noisier \texttt{MMR}.

Finally, we observe similar trends across models, indicating that \texttt{MMR} is robust to the choice of ASR model. The pattern holds even for MMS, an encoder-only CTC model, suggesting that the effect does not stem primarily from language modeling. For each language, \Cref{fig:partial-spearman} shows the partial correlation between \texttt{FL} and \texttt{MMR}, using the model whose \texttt{MMR} correlates most strongly with \texttt{FL}.


\section{Analysis}\label{sec:analysis}
\begin{figure}[t!]
    \centering
    \includegraphics[width=0.9\linewidth]{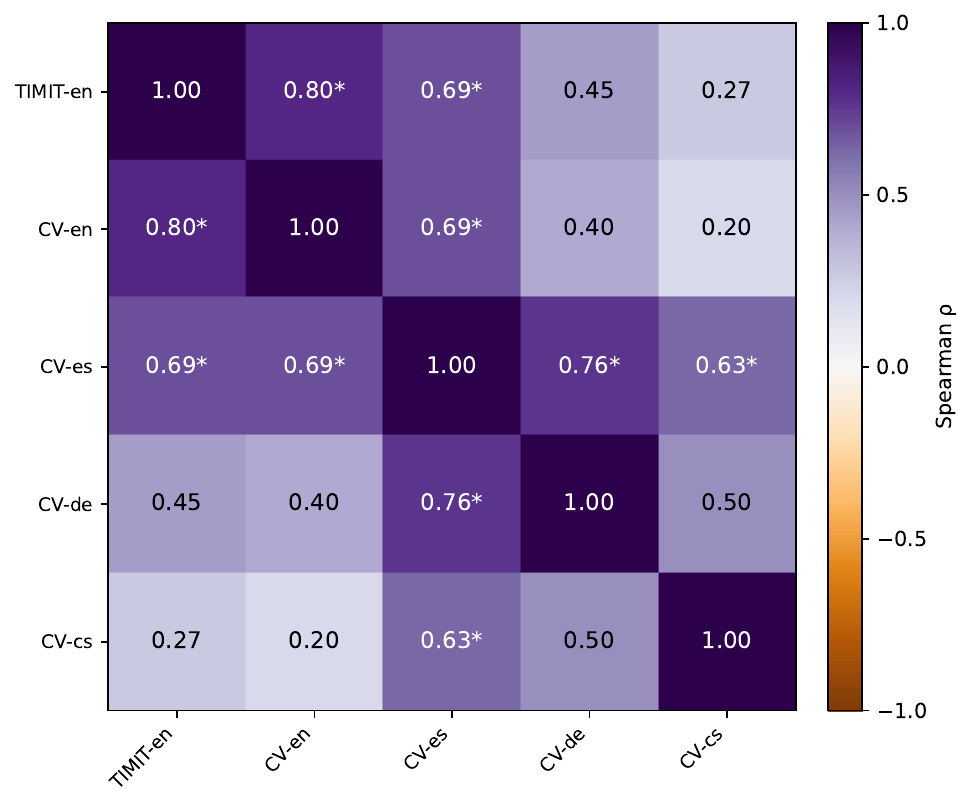}
\caption{Pairwise Spearman rank correlation of \texttt{MMR}, using the same per-language ASR models as \Cref{fig:partial-spearman}. Each cell reports $\rho$ between the two datasets' \texttt{MMR} rankings of the shared consonants. $^{*}p<.05$.}
\label{fig:cross-lang}
\end{figure} 

\subsection{Cross-language analysis on consonant contribution}\label{ssec:cross-analysis} 
Consonant contribution depends on each language's phonology, so consonant rankings need not agree across languages. We therefore compare \texttt{MMR} rankings between corpora. \Cref{fig:cross-lang} shows pairwise Spearman rank correlations of \texttt{MMR} over the 12 consonants shared by all datasets (\textipa{b, d, g, p, t, k, l, m, n, f, s, S}). The two English corpora (TIMIT and CV-en) agree most strongly ($\rho=0.80^{*}$). Spanish (CV-es) shows the most consistent cross-language agreement, correlating significantly with every other corpus ($\rho=0.63$--$0.76$, all $p<.05$). This may reflect its small, cross-linguistically common consonant inventory overlapping with each of the other languages. By contrast, the two English corpora show no significant agreement with German (CV-de) or Czech (CV-cs), and German and Czech also fail to reach significance ($\rho=0.50$, $p=.10$). Overall, consonant contribution rankings hold only partially across languages, indicating that consonant contribution is language-specific rather than language-universal.

\subsection{Positional asymmetry in consonant contribution}\label{ssec:initial-final}
\begin{figure}[t!]
    \centering
    \includegraphics[width=0.9\linewidth]{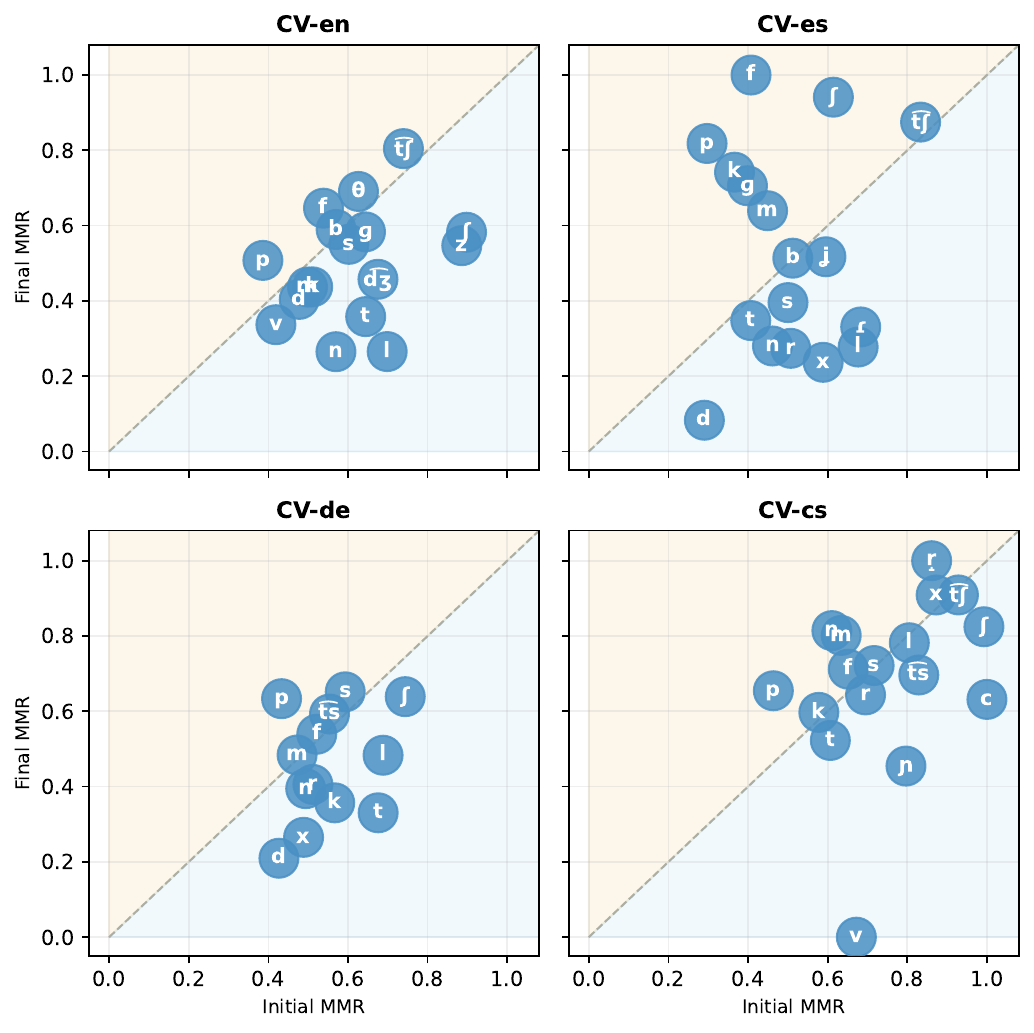}
\caption{Per-consonant \texttt{MMR} in word-initial (x-axis) versus word-final (y-axis) across four languages (Whisper). Each point represents a consonant occurring in both word-initial and word-final positions ($n \geq 30$ observations per position). Points above the diagonal (orange) have higher final than initial \texttt{MMR}, while points below (blue) have the reverse.}
\label{fig:position}
\end{figure}

Consonant contribution may also depend on word position, so we compare \texttt{MMR} for word-initial and word-final consonants (\Cref{fig:position}).
In English and German, masking word-initial consonants is generally more disruptive than masking word-final consonants, consistent with the perceptual primacy of word onsets in lexical access \cite{marslenwilson1978,marslenwilson1989,gurevichkim2024}.
However, this onset advantage is not consistent across consonants. Some consonants show comparable or greater disruption in final position, and Spanish and Czech exhibit no clear positional trend. Onset prominence is therefore a useful but limited signal, and future rankings could account for the positional effect differing across consonants rather than assuming a uniform onset advantage. Confirming these patterns will require further investigation across datasets within each language.

\section{Discussion}\label{sec:discussion}
\subsection{Integration with FITI}
FITI \cite{gurevich2023,gurevichkim2024} includes both \texttt{freq} and \texttt{FL} as linguistic factors for consonant contribution to intelligibility, but does not isolate their separate contributions despite their strong correlation (\Cref{fig:collinearity}).
After separating them with partial correlations, we find the two behave differently.
Higher \texttt{FL} predicts greater masking-induced disruption, aligning with FITI's claim that phonemes carrying more lexical contrast matter more to intelligibility.
\texttt{freq}, on the other hand, seems to show the opposite pattern: more frequent consonants are less disruptive under masking.
We interpret this inversion in information-theoretic terms: frequent consonants tend to be predictable from the contexts in which they occur \cite{cohenpriva2015informativity}, so the word often remains recoverable from the unmasked segments. This pattern is unlikely to be solely an ASR artifact: if recognizer priors were the main cause, the negative frequency effect should strengthen from encoder-only MMS to encoder-decoder Whisper and LLM-based Qwen3-ASR, but \Cref{tab:fl_corr} shows similar results.

The opposing signs mean the two factors encode different notions of contribution rather than a single scale. \texttt{FL} reflects per-instance necessity, marking consonants whose loss makes the containing word hard to recognize, while \texttt{freq} reflects lexical coverage, marking phonemes that recur across more of the vocabulary. Because a phoneme can rank high on one and low on the other, the choice between them follows from the clinical objective: expanding the number of words a speaker produces intelligibly favors \texttt{freq}, whereas improving the recognizability of high-impact consonants favors \texttt{FL}.

\subsection{Clinical considerations}
Although \texttt{MMR} yields a language-level prior over intervention targets, target selection ultimately remains individual: the ranking is a starting point that speaker-specific factors can override. One possible factor is \textit{stimulability}. As dysarthria constrains the articulatory space unevenly, a consonant ranked high by \texttt{MMR} may lie outside a speaker's current motor range, while a lower-ranked one may be more attainable. Another factor can be \textit{communicative centrality}. Corpus-level statistics reflect the frequency and contrastiveness of words in the language at large, but omit content specific to the individual, such as family names, place names, and vocabulary tied to daily routines. The phonemes these words depend on may carry more weight for a given speaker than any language-level ranking implies, shifting priority toward sounds the corpus would rank lower.

\section{Conclusion}\label{sec:conclusion}
We presented a scalable method for quantifying consonant contribution to intelligibility across English, Spanish, German, and Czech using acoustic masking paired with ASR. By silencing one consonant at a time and measuring whether the affected word remains recognizable, we derived empirical consonant contribution rankings (\texttt{MMR}) and related them to known linguistic factors (\texttt{freq} and \texttt{FL}). These rankings can offer a language-level prior for prioritizing treatment targets in motor speech disorders.

Several limitations bound these findings. First, acoustic masking via silence is not equivalent to real-world mispronunciation: a dysarthric speaker typically preserves partial cues such as manner of articulation or spectral shape, whereas silence removes everything. Future work could address this by modeling graded, feature-specific distortions using phonological vector steering applied on self-supervised representations \cite{choi2026b}. Second, we mask one consonant at a time, so the measure isolates each consonant's individual contribution and does not capture phonotactic interactions between adjacent segments, such as within consonant clusters. Finally, since ASR misrecognition is a proxy for human intelligibility, perceptual validation with listeners is still needed to confirm that the rankings reflect human judgments.



 


\section*{Acknowledgements}
This work was supported in part by Texas Health Catalyst, a program of the Office of Innovation and Entrepreneurship at Dell Medical School at The University of Texas at Austin. The authors also acknowledge the program’s mentors and advisors for their guidance and commercialization support.

\section*{AI-Generated Content Disclosure}
Generative artificial intelligence tools were used to assist with language editing, clarity of presentation, and code drafting/debugging. All research ideas, methodology, experiments, analyses, and interpretations were conceived, verified, and carried out by the authors, who take full responsibility for the originality, validity, and integrity of the work.

\bibliographystyle{IEEEtran}
\bibliography{mybib}

\end{document}